\documentclass[letterpaper]{article}
\usepackage{aaai}  
\usepackage{times}  
\usepackage{helvet}  
\usepackage{courier}  
\usepackage[hyphens]{url}  
\usepackage{graphicx} 
\usepackage{natbib}  
\usepackage{caption} 
\usepackage{multirow}
\usepackage{amsmath}
\usepackage{amssymb}
\usepackage{hyperref}
\usepackage{cleveref}
\usepackage{booktabs}
\usepackage{multirow}
\usepackage{graphicx}
\usepackage{makecell}
\copyrighttext{}
 
\begin{document}
%
\title{HT-Bench: Benchmarking and Learning Dexterous Full-Hand Tactile Representations with Egocentric Vision}

\author{
    Yuzhe Huang\textsuperscript{\rm 1,2$\star$}, 
    Jiaping Wu\textsuperscript{\rm 2,3$\star$}, 
    Jiaming Jiang\textsuperscript{\rm 4}, 
    Hezhe Lin\textsuperscript{\rm 2,5}, 
    Aikebaier Aierken\textsuperscript{\rm 2,6}, \\
    \Large\bf{Yunlong Wang}\textsuperscript{\rm 2}, 
    Kun Cheng\textsuperscript{\rm 3},
    Wanlin Li\textsuperscript{\rm 7},
    Chenxi Xiao\textsuperscript{\rm 4},
    Ziyuan Jiao\textsuperscript{\rm 1$\dagger$}, 
    Yuanxin Zhong\textsuperscript{\rm 2$\dagger,\ddagger$} \\
    \textsuperscript{\rm 1}Beihang University \quad
    \textsuperscript{\rm 2}Rimbot \quad
    \textsuperscript{\rm 3}BUPT \quad
    \textsuperscript{\rm 4}ShanghaiTech University \quad
    \textsuperscript{\rm 5}Tsinghua University \quad
    \textsuperscript{\rm 6}CAS \quad
    \textsuperscript{\rm 6}Bigai \\
    \textsuperscript{$\star$}Equal contributors \quad 
    \textsuperscript{$\dagger$}Corresponding authors \quad 
    \textsuperscript{$\ddagger$}Project Lead
}

\maketitle
\begin{abstract}
\begin{quote}

Establishing a universal benchmark for tactile representation learning in robotic manipulation remains challenging due to the diversity of tactile sensor designs, data formats, and robot embodiments.
Rather than seeking to establish such, we explore a scalable and promising direction for future development: egocentric vision paired with full-hand tactile data.
To this end, we introduce \textbf{HT-Bench}, a large-scale multi-task benchmark containing 10M RGB frames and 7.8M tactile frames across 226 tasks. 
HT-Bench evaluates contact-geometry modeling, vision--touch alignment, and generalization to unseen tasks through fine-grained tactile retrieval, masked tactile inpainting, vision-to-tactile synthesis, and multimodal tactile frame prediction. 
We further propose \textbf{HandTouch}, a vector-quantized vision--tactile encoder trained progressively with spatial, cross-modal, and temporal objectives. 
HandTouch consistently outperforms representative baselines on HT-Bench, improving Recall@5 from 74.65\% to 85.23\%, reducing masked inpainting RMSE from 0.022 to 0.009, and increasing vision-to-tactile test cIoU from 0.628 to 0.689. 
Moreover, in real-world downstream manipulation, HandTouch achieves the best average success rate of 68.3\% across four contact-rich tasks, outperforming the strongest baseline by 18.3 percentage points. 
These results show that HandTouch learns transferable tactile representations and that egocentric full-hand tactile data offers a scalable basis for dexterous tactile learning.
\end{quote}
\end{abstract}

\begin{figure*}[ht!]
    \centering
    \includegraphics[width=\linewidth]{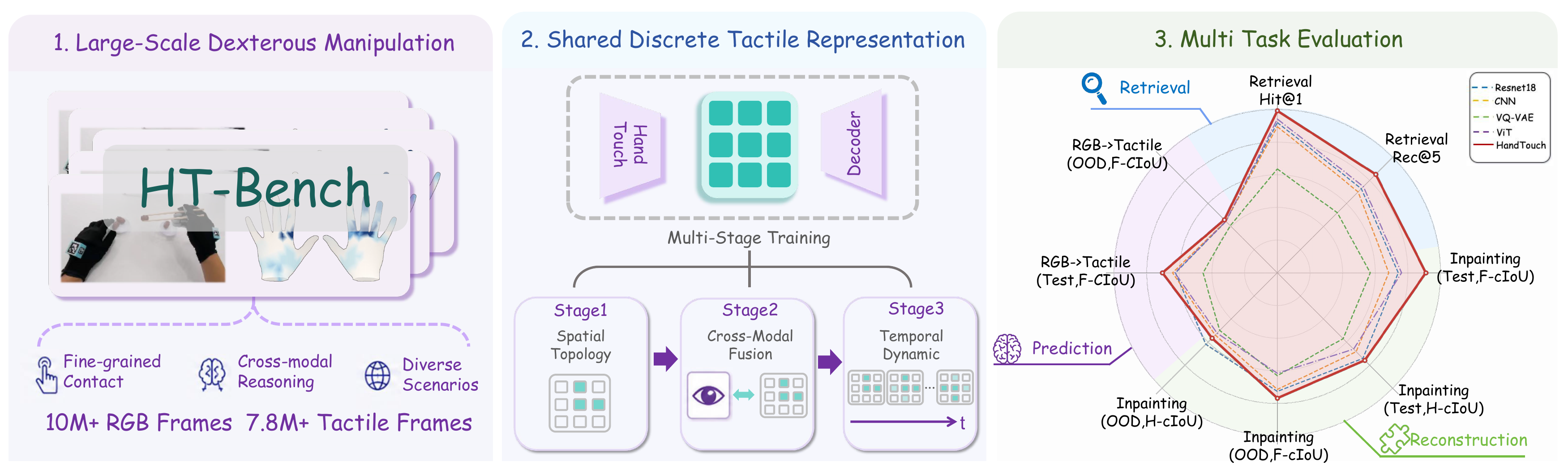}
    \caption{Overview of HT-Bench. 1. HT-Bench pairs egocentric vision with full-hand tactile data to provide a scalable benchmark for dexterous tactile representation learning. It contains 10M RGB frames and 7.8M tactile frames collected from diverse manipulation tasks. 2. HandTouch learns a shared discrete tactile representation through progressive spatial, cross-modal, and temporal training, and 3. is evaluated on fine-grained tactile similarity retrieval, masked tactile inpainting, vision-to-tactile synthesis, and multimodal tactile frame prediction under task-level out-of-distribution splits.}
    \label{fig:teaser}
\end{figure*}

\section{Introduction}
Tactile sensing has attracted increasing attention as a key modality for building robust multimodal foundation models for robotic manipulation, as it captures direct physical interactions---including contact forces, pressure distributions, and slip detection---that vision alone cannot reliably estimate~\citep{yang2024bindingtouch,luo2026omniumi,huang2026tactile}. With the paradigm shift from vision-centric perception to multimodal robotic policies, tactile sensing has become increasingly important for systems requiring contact-rich reasoning~\citep{chen2026multimodal,li2026simultaneous} and precise dexterous manipulation~\citep{lin2025pp,DexMove_ICLR2026}. Yet, unlike visual perception, which benefits from relatively standardized data formats, scalable architectures, and mature evaluation benchmarks~\citep{dosovitskiy2021vit,radford2021clip,zhai2023sigmoid,russakovsky2015imagenet,lin2015microsoftcoco}, tactile representation learning remains constrained by heterogeneous datasets, sensor-specific processing pipelines, and task-dependent evaluation protocols~\citep{chen2026univtac,cao2026tactile}.

This challenge stems from the intrinsic heterogeneity of tactile information across sensor designs and robotic embodiments. Tactile sensors vary in hardware principles, spatial layouts, signal modalities, and mounting configurations, while robot embodiments introduce different end-effector morphologies and contact patterns~\citep{schneider2025tactile,cao2026tactile,chen2026univtac}. Thus, building a universal benchmark that fully accommodates these discrepancies remains impractical. Instead, we focus on a scalable direction: pairing egocentric vision with full-hand tactile data. Egocentric observations provide interaction-centric visual context, while full-hand tactile sensing captures distributed contact during dexterous manipulation. However, existing evaluations still leave open a key question: what encoder can serve as an effective representation backbone for dexterous full-hand tactile perception?

To address this gap, we introduce \textbf{HT-Bench} (as shown in \cref{fig:teaser}left), a large-scale multi-task benchmark for evaluating tactile representation learning in dexterous full-hand sensing. HT-Bench aggregates synchronized egocentric visual observations and full-hand tactile sequences, and evaluates encoders under four tasks: fine-grained tactile similarity retrieval, masked tactile inpainting, vision-to-tactile synthesis, and multimodal tactile frame prediction. Instead of relying on a single downstream objective, HT-Bench examines tactile representations from three perspectives: whether they encode meaningful contact geometry, whether they align tactile observations with visual information, and whether they generalize to unseen interaction tasks.

Building on HT-Bench, we propose \textbf{HandTouch}, a vector-quantized vision--tactile encoder for general dexterous tactile representation learning. As shown in \cref{fig:teaser}middle, HandTouch follows a progressive training pipeline: it first captures tactile spatial topology via vector-quantized reconstruction, mapping continuous tactile observations into a shared discrete token space; then learns vision--tactile alignment through cross-modal masked tactile inpainting; and finally models temporal contact evolution with multimodal tactile frame prediction. 


Experiments on HT-Bench show that HandTouch achieves stronger performance than representative tactile encoder baselines. On fine-grained tactile similarity retrieval, HandTouch improves Recall@5 from 74.65\% to 85.23\% compared with the strongest baseline. On masked tactile inpainting, it reduces full-image RMSE from 0.022 to 0.009 and improves full-image cIoU from 0.762 to 0.912 on the standard test split. For vision-to-tactile synthesis, HandTouch improves OOD cIoU from 0.628 to 0.689, indicating stronger cross-modal tactile generation and better generalization to unseen tasks. Moreover, in real-world downstream manipulation, HandTouch achieves the best average success rate of 68.3\% across four contact-rich tasks, outperforming the strongest baseline by 18.3 percentage points. 


In summary, our contributions are as follows:
\begin{itemize}
\item We introduce \textbf{HT-Bench}, a large-scale multi-task benchmark for evaluating dexterous full-hand tactile representation learning based on egocentric vision paired with full-hand tactile data.

\item We propose \textbf{HandTouch}, a vector-quantized vision--tactile encoder that learns tactile representations through progressive spatial, cross-modal, and temporal training.

\item HandTouch achieves consistently stronger performance than representative tactile encoder baselines on HT-Bench and downstream task, demonstrating improved structural retrieval, masked reconstruction, cross-modal synthesis, and out-of-distribution generalization.
\end{itemize}

We will release all data, evaluation protocols, pretrained weights, and training/testing scripts to facilitate future research in general tactile representation learning.

\begin{figure*}[t!]
    \centering
    \includegraphics[width=\linewidth]{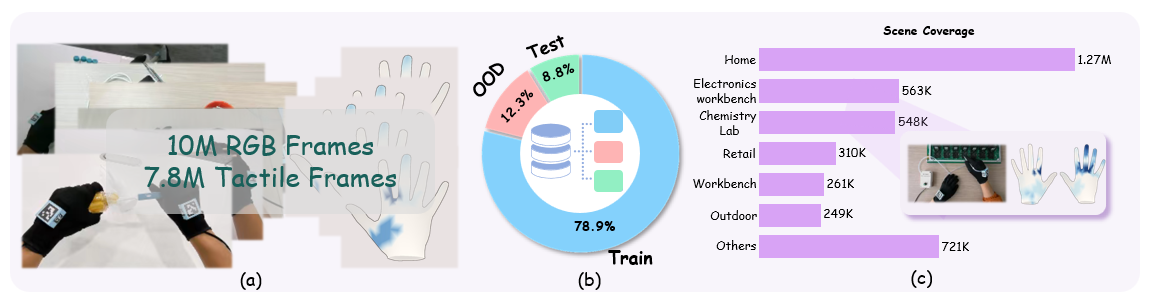}
    \caption{\textbf{Statistics and coverage of HT-Bench.}
    HT-Bench contains 10M RGB frames and 7.8M full-hand tactile frames from dexterous manipulation, with training, test, and task-level OOD splits. It covers diverse scenes such as home, electronics workbench, chemistry lab, retail, workbench, outdoor, and other environments.}
    \label{fig:HT-bench}
    \vspace{-4pt}
\end{figure*}

\begin{table*}[t!]
\centering
\renewcommand{\arraystretch}{0.9}
\setlength{\tabcolsep}{16pt}
\caption{Comparison with existing tactile benchmarks. HT-Bench jointly supports dexterous full-hand sensing, multi-task evaluation, multi-scene data, and task-level OOD splits.}
\label{tab:benchmark_comparison}
\begin{tabular}{lcccc}
\toprule
\textbf{Benchmark} & \textbf{Full-Hand} & \textbf{Tasks} & \textbf{Scenes} & \textbf{OOD Split} \\
\midrule
Sparsh~\citep{higuera2024sparsh} & $\times$ & Multi-task & Single-scene & $\times$ \\
AnyTouch 2~\citep{feng2026anytouch} & $\times$ & Multi-task & Multi-scene & $\times$ \\
OpenTouch~\citep{song2025opentouch} & $\checkmark$ & Multi-task & Multi-scene & $\times$ \\
TouchAnything~\citep{zhou2026touchanything} & $\checkmark$ & Single-task & Multi-scene & $\checkmark$ \\
\textbf{HT-Bench (Ours)} & \textbf{\checkmark} & \textbf{Multi-task} & \textbf{Multi-scene} & \textbf{\checkmark} \\
\bottomrule
\end{tabular}
\vspace{-8pt}
\end{table*}

\section{Related Work}
\noindent\textbf{Tactile and visuo-tactile representation learning.}
Tactile representation learning has been studied for object recognition~\citep{yang2024binding,huang2026tactile}, contact understanding~\citep{xie2025universal}, manipulation feedback~\citep{li2025visuo}, and visuo-tactile prediction~\citep{chen2026univtac}. 
Recent methods show that tactile signals provide complementary physical cues for contact-rich dexterous manipulation. 
However, most approaches learn representations in task-specific pipelines, where features are tightly coupled with particular sensor configurations, interaction domains, or downstream objectives~\citep{liu2024masked,zorin2026taco}. 
Thus, it remains unclear whether these features are transferable tactile representations or task-specific adaptations. 
This motivates HT-Bench, which evaluates tactile encoders beyond a single objective by assessing contact geometry modeling, visual alignment, temporal contact dynamics, and generalization to unseen tasks. 
Based on this framework, HandTouch learns general dexterous tactile representations through progressive spatial, cross-modal, and temporal objectives.


\noindent\textbf{Tactile datasets and benchmarks.}
Recent tactile datasets and benchmarks have advanced tactile learning but also reveal its intrinsic heterogeneity across sensors, embodiments, and tasks. Resources such as Sparsh~\citep{higuera2024sparsh}, AnyTouch~\citep{feng2025anytouch,feng2026anytouch}, OpenTouch~\citep{song2025opentouch}, and TouchAnything~\citep{zhou2026touchanything} support tactile perception, visuo-tactile learning, or touch-centric manipulation across diverse platforms and scenarios. Other datasets, including VT-DexManip~\citep{liu2025vtdexmanip}, multi-modal tactile datasets~\citep{chi2024multi}, and UniVTac~\citep{chen2026univtac}, further study dexterous manipulation, synchronized sensory streams, or unified visuo-tactile modeling under specific settings. However, variations in sensing principles, spatial layouts, signal formats, and robot embodiments make a universal benchmark difficult to define.

HT-Bench therefore does not aim to resolve all tactile heterogeneity. Instead, we focus on a scalable regime for future embodied AI~\citep{zheng2026egoscalescalingdexterousmanipulation,humanego2026}: egocentric vision paired with full-hand tactile sensing. Egocentric observations provide hand-centric interaction context~\citep{fang2026unidextok}, while full-hand tactile sensing captures distributed contact patterns in dexterous manipulation. This pairing enables evaluation of contact-geometry modeling, visual alignment, and unseen-task generalization. HT-Bench complements existing datasets with a standardized four-track multi-task protocol in this setting, as compared in \cref{tab:benchmark_comparison}.


\noindent\textbf{Masked and discrete representation learning.}
Masked modeling and vector-quantized representation learning are effective for compact visual and multimodal representations~\citep{razavi2019generating,bao2021beit,van2017neural}. 
However, applying them to dexterous hand tactile signals is challenging because tactile observations are sparse, locally structured, and coupled with interaction dynamics~\citep{wu2026dexgrasp,xie2026universal}. 
HandTouch adapts these ideas to full-hand tactile learning through three progressive stages: vector-quantized reconstruction for spatial topology, cross-modal masked inpainting for vision--touch alignment, and multimodal tactile frame prediction for temporal reasoning. 
These objectives encourage structurally discriminative and visually grounded tactile representations aligned with the core capabilities evaluated by HT-Bench.

\section{HT-Bench: A Multi-Task Tactile Evaluation Benchmark}
\label{sec:ht_bench}

Extending existing open-source tactile and visuo-tactile datasets~\citep{song2025opentouch,zhou2026touchanything} with newly collected real-world full-hand tactile sequences, we construct \textbf{HT-Bench} for evaluating dexterous full-hand tactile representation learning. HT-Bench contains approximately 10M RGB frames and 7.8M tactile frames, as shown in \cref{fig:HT-bench}(a). To assess out-of-distribution (OOD) generalization, we adopt a task-level split: one interaction task is held out for OOD evaluation, while the remaining tasks are divided into training and in-distribution test sets with a 9:1 ratio; see \cref{fig:HT-bench}(b). The data volume and episode distribution across fine-grained scene categories are summarized in \cref{fig:HT-bench}(c). Based on this dataset, we define four tasks that evaluate complementary capabilities: contact structure understanding, vision--touch alignment, spatial reasoning, and temporal contact modeling.

\noindent\textbf{Fine-Grained Tactile Similarity Retrieval.}
We evaluate whether tactile embeddings preserve fine-grained contact characteristics with a 1-vs-20 retrieval task. For each query tactile map, candidates are ranked by Structural Similarity Index Measure (SSIM) to obtain tactile-structure reference rankings, which are compared with cosine-similarity rankings induced by learned embeddings.

\noindent\textbf{Masked Tactile Inpainting.}
We evaluate spatial reasoning under incomplete tactile observations by masking regions of the tactile map and requiring the model to reconstruct missing tactile responses from remaining tactile signals and visual cues. This setting mimics partial tactile information loss from sensor failures or degraded regions, and tests whether representations capture spatially coherent contact patterns.

\noindent\textbf{Vision-to-Tactile Synthesis (RGB-to-Tactile).}
We evaluate vision--touch alignment by predicting tactile pressure distributions from a single RGB observation. Motivated by humans' ability to form tactile expectations from visual perception, related to cross-modal sensory integration and synesthetic associations~\citep{cytowic2002synesthesia}, this task examines whether learned representations capture the relationship between visual observations and tactile responses induced by object geometry and physical interaction.

\noindent\textbf{Multimodal Tactile Frame Prediction.}
We evaluate temporal contact modeling by predicting the tactile distribution $\mathbf{t}_T$ from recent visual observations $\mathbf{v}_{T-2}$ and past tactile trajectories $\mathbf{t}_{T-2}$. This task tests whether the representation can integrate visual context and tactile history to anticipate future contact states and capture the temporal evolution of tactile interactions.

\section{HandTouch: Vector-Quantized Vision--Tactile Representation Learning}
\label{sec:HandTouch}





Given the evaluation requirements defined by HT-Bench, we propose \textbf{HandTouch}, a vector-quantized vision--tactile encoder for learning transferable full-hand tactile representations. The overall pipeline is illustrated in \cref{fig:pipeline}. HandTouch is trained progressively through three stages, each corresponding to a core capability evaluated by HT-Bench. First, it learns unimodal tactile reconstruction to capture spatial contact topology (\cref{sec:train-stage1}). Second, it incorporates visual cues through cross-modal masked tactile inpainting to learn vision-tactile alignment (\cref{sec:train-stage2}). Finally, it models contact dynamics through multimodal tactile frame prediction (\cref{sec:train-stage3}).

\subsection{Vector-Quantized Tactile Reconstruction}\label{sec:train-stage1}
\begin{figure*}[ht!]
    \centering
    \includegraphics[width=\linewidth]{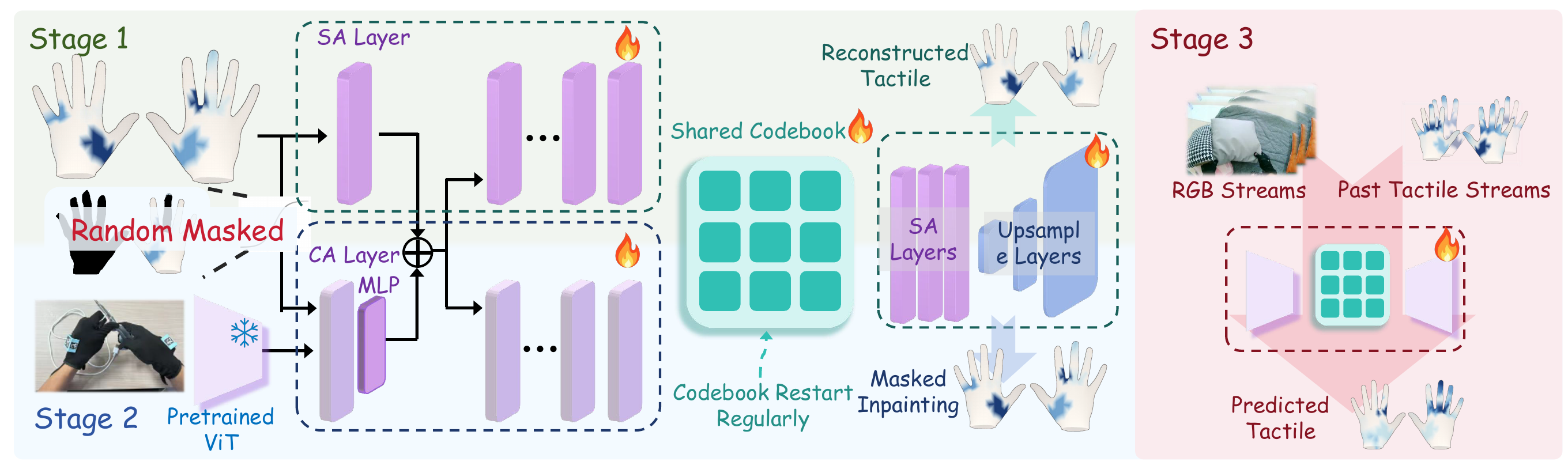}
    \caption{\textbf{Training pipeline of HandTouch.} 
    \textbf{Stage 1:} Learning spatial topologies of tactile graphics via unimodal self-attention reconstruction and vector quantization with a shared codebook. 
    \textbf{Stage 2:} Reconstructing highly corrupted tactile images under a dynamic regional/complete masking scheme, guided by visual priors injected through cross-attention. 
    \textbf{Stage 3:} Forecasting the current tactile distribution $\mathbf{t}_T$ based on sequential visual context $\mathbf{v}_{T-2:T}$ and past tactile histories $\mathbf{t}_{T-2:T-1}$. Modules with flame icons are actively trained in each phase.}
    \label{fig:pipeline}
\end{figure*}

As illustrated in \cref{fig:pipeline} (upper-left), Stage 1 learns a unimodal tactile reconstruction objective to capture the spatial topology of full-hand tactile maps. Given a normalized tactile map $\mathbf{t} \in [0,1]^{1 \times 224 \times 224}$, a convolutional projection layer tokenizes it into non-overlapping patches. After adding learnable positional embeddings, the patch tokens are processed by an 8-layer Vision Transformer (ViT) encoder, producing continuous latent features $\mathbf{Z}_e \in \mathbb{R}^{N \times D}$.

To construct a compact and robust discrete feature space, we employ a factorized vector quantizer. Let $\mathcal{C} = \{\mathbf{e}_i\}_{i=1}^K \subset \mathbb{R}^{d}$ denote the shared codebook of size $K=2048$ with a lower bottleneck dimension $d \ll D$. For the $j$-th patch embedding $\mathbf{z}_e^{(j)} \in \mathbb{R}^D$, we first project it into the codebook space using an input projection $\mathbf{W}_{\text{in}} \in \mathbb{R}^{d \times D}$:
\begin{equation}
\mathbf{z}_q^{(j)} = \mathbf{e}_k, \quad \text{where } k = \arg\min_{i} \|\mathbf{W}_{\text{in}}\mathbf{z}_e^{(j)} - \mathbf{e}_i\|_2^2
\label{eq:quantization}
\end{equation}
The quantized token is mapped back to the encoder embedding space by an output projection $\mathbf{W}_{\text{out}} \in \mathbb{R}^{D \times d}$ before being passed to the decoder.

A common issue in vector quantization is codebook collapse, where only a small subset of codebook entries is frequently selected while many entries remain inactive. To mitigate this issue, we track the usage frequency of each codebook entry with an exponential moving average during training. Codebook entries whose cumulative usage falls below a restart threshold $\tau$ are reinitialized using randomly sampled active projected features from $\mathbf{W}_{\mathrm{in}}\mathbf{Z}_e$ in the current batch, with small isotropic Gaussian noise added for exploration.

The decoder reconstructs the input tactile map $\hat{\mathbf{t}}$ from the quantized tokens using attention blocks and convolutional upsampling. The Stage 1 objective is defined as:
\begin{equation}
\begin{split}
\mathcal{L}_{\text{stage1}} &= \|\mathbf{t} - \hat{\mathbf{t}}\|^2_2 + \|\mathbf{Z}_q - \operatorname{sg}[\mathbf{W}_{\text{in}}\mathbf{Z}_e]\|_2^2 \\
&\quad + \beta \|\operatorname{sg}[\mathbf{Z}_q] - \mathbf{W}_{\text{in}}\mathbf{Z}_e\|_2^2
\end{split}
\label{eq:stage1_loss}
\end{equation}
where $\operatorname{sg}[\cdot]$ denotes the stop-gradient operator and $\beta$ is the commitment loss weight. All modules in this stage are optimized jointly, as indicated by the flame icons in \cref{fig:pipeline}.

\subsection{Cross-Modal Masked Tactile Inpainting}\label{sec:train-stage2}

Stage 2 extends HandTouch from unimodal tactile reconstruction to cross-modal vision--tactile alignment through \textit{Cross-Modal Masked Tactile Inpainting}. As illustrated in \cref{fig:pipeline} (bottom-left), the objective is to reconstruct corrupted tactile maps by using both the remaining tactile observations and synchronized visual cues.

Given a tactile map $\mathbf{t}$, we generate a masked tactile input $\tilde{\mathbf{t}}$ using a curriculum-based dual masking strategy. The first masking mode is \textit{Regional Random Masking}, where tactile regions corresponding to anatomical hand parts, such as the thumb, index finger, middle finger, or palm, are masked to simulate partial tactile information loss. The second mode is \textit{Complete Masking}, where the tactile map is entirely masked, and the model must infer the tactile response primarily from visual context. To gradually increase task difficulty, the probability of applying Complete Masking increases over training. Let $\gamma \in [0,1]$ denote the normalized training progress. We define:
\begin{equation}
P_{\mathrm{full}}(\gamma)
= p_{\min} + \frac{(p_{\max}-p_{\min})}{1+\exp[-12(\gamma-0.5)]},
\label{eq:dynamic_mask}
\end{equation}
where $p_{\min}$ and $p_{\max}$ denote the minimum and maximum probabilities of Complete Masking, respectively. The remaining probability, $1-P_{\mathrm{full}}(\gamma)$, is assigned to Regional Random Masking. This curriculum gradually shifts the training objective from local tactile inpainting to more challenging vision-conditioned tactile synthesis.

The synchronized visual frame $\mathbf{v}$ is encoded by a frozen pre-trained ViT to extract visual context features $\mathbf{F}_v$. To inject visual information into the tactile reconstruction process, we apply a cross-attention layer followed by a multi-layer perceptron. The tactile tokens from the masked tactile input serve as queries, while visual features serve as keys and values. This design allows the tactile decoder to use visual context when reconstructing missing tactile regions. Since Stage 2 processes tactile observations from both left and right hands, we additionally introduce a learnable hand-specific token $\mathbf{t}_{\mathrm{hand}} \in \{\mathbf{t}_{\mathrm{left}}, \mathbf{t}_{\mathrm{right}}\}$ to reduce geometric ambiguity caused by lateral mirroring.

The Stage 2 loss combines reconstruction terms with the vector-quantization losses inherited from Stage 1. Let $\mathbf{M} \in \{0, 1\}^{1 \times 224 \times 224}$ denote a binary occlusion mask, where $\mathbf{M}_{c,h,w}=1$ indicates a masked pixel.
\begin{equation}
\begin{split}
\mathcal{L}_{\text{stage2}} 
&= \lambda_{\text{vis}} \|(\mathbf{1}-\mathbf{M}) \odot (\mathbf{t} - \hat{\mathbf{t}}_{\text{cm}})\|^2_2 \\
&\quad + \lambda_{\text{mask}} \|\mathbf{M} \odot (\mathbf{t} - \hat{\mathbf{t}}_{\text{cm}})\|^2_2 \\
&\quad + \|\mathbf{Z}_q - \operatorname{sg}[\mathbf{W}_{\text{in}}\mathbf{Z}_e]\|_2^2 \\
&\quad + \beta \|\operatorname{sg}[\mathbf{Z}_q] - \mathbf{W}_{\text{in}}\mathbf{Z}_e\|_2^2
\end{split}
\label{eq:stage2_loss}
\end{equation}
where $\odot$ denotes the Hadamard product, $\hat{\mathbf{t}}_{\mathrm{cm}}$ is the cross-modally reconstructed tactile output, and $\lambda_{\mathrm{vis}}$ and $\lambda_{\mathrm{mask}}$ balance the visible-region and masked-region reconstruction losses. We set $\lambda_{\mathrm{mask}} > \lambda_{\mathrm{vis}}$ to emphasize reconstruction quality in the missing tactile regions. The codebook restart mechanism from \cref{sec:train-stage1} remains active in this stage to prevent codebook underutilization. All non-frozen modules are optimized jointly.

\subsection{Multimodal Tactile Frame Prediction}\label{sec:train-stage3}
In the final stage, HandTouch learns temporal contact modeling through \textit{Multimodal Tactile Frame Prediction}. As illustrated in \cref{fig:pipeline} (right), this stage predicts the tactile frame at time step $T$ by integrating recent visual observations and past tactile history.

Formally, the model takes recent visual observations $\mathbf{v}_{T-2}$ and past tactile trajectories $\mathbf{t}_{T-2}$ as input. These multimodal sequences are temporally aggregated and projected into the shared discrete latent space learned in the previous stages. The decoder then predicts the tactile distribution $\hat{\mathbf{t}}_T$ at time step $T$. This objective requires the model to combine visual context with tactile history, thereby encouraging temporally aware tactile representations.

The Stage 3 objective is defined as:
\begin{equation}
\mathcal{L}_{\mathrm{stage3}} = |\mathbf{t}_T-\hat{\mathbf{t}}_T|_2^2 .
\label{eq}
\end{equation}
During this stage, the prediction modules, shared codebook, and decoder are fine-tuned to improve multimodal tactile prediction, as indicated by the flame icons in \cref{fig:pipeline}.

\begin{table*}[t]
\centering
\caption{\textbf{Quantitative comparison on HT-Bench.} We compare HandTouch with representative tactile encoder baselines on fine-grained tactile similarity retrieval, masked tactile inpainting, and vision-to-tactile synthesis (RGB$\to$Tac). Retrieval performance is evaluated using Hit@1 and Recall@5, while inpainting and synthesis are evaluated on the standard test and task-level OOD splits using RMSE and contact IoU (cIoU). `F-' and `H-' denote metrics computed over the full tactile map and masked hole regions, respectively. $\uparrow$ and $\downarrow$ indicate that higher and lower is better, and the best results are highlighted in \textbf{bold}.}
\label{tab:main_results}
\resizebox{\textwidth}{!}{%
\begin{tabular}{l c c c c c c c c c c c c c c}
\toprule
\multirow{2}{*}{\textbf{Model}} & \multicolumn{2}{c}{\textbf{Retrieval}} & \multicolumn{4}{c}{\textbf{Inpainting (Test)}} & \multicolumn{4}{c}{\textbf{Inpainting (OOD)}} & \multicolumn{2}{c}{\textbf{RGB$\to$Tac (Test)}} & \multicolumn{2}{c}{\textbf{RGB$\to$Tac (OOD)}} \\
\cmidrule(lr){2-3} \cmidrule(lr){4-7} \cmidrule(lr){8-11} \cmidrule(lr){12-13} \cmidrule(lr){14-15}
& Hit@1$\uparrow$ & Rec@5$\uparrow$ & F-RMSE$\downarrow$ & F-cIoU$\uparrow$ & H-RMSE$\downarrow$ & H-cIoU$\uparrow$ & F-RMSE$\downarrow$ & F-cIoU$\uparrow$ & H-RMSE$\downarrow$ & H-cIoU$\uparrow$ & F-RMSE$\downarrow$ & F-cIoU$\uparrow$ & F-RMSE$\downarrow$ & F-cIoU$\uparrow$ \\
\midrule
ResNet-18 & 92.13 & 72.85 & 0.025 & 0.742 & 0.025 & 0.742 & 0.041 & 0.727 & \textbf{0.056} & \textbf{0.620} & 0.038 & 0.624 & 0.084 & 0.445 \\
CNN-based & 89.61 & 70.09 & 0.030 & 0.684 & 0.030 & 0.684 & 0.047 & 0.715 & 0.068 & 0.538 & 0.036 & 0.642 & 0.083 & 0.447 \\
VQ-VAE    & 63.60 & 51.98 & 0.042 & 0.570 & 0.042 & 0.570 & 0.053 & 0.630 & 0.068 & 0.499 & 0.060 & 0.456 & 0.081 & 0.408 \\
ViT-based & 94.27 & 74.65 & 0.022 & 0.762 & 0.033 & 0.662 & 0.056 & 0.615 & 0.065 & 0.522 & 0.038 & 0.628 & 0.083 & 0.446 \\
\textbf{Ours} & \textbf{99.27} & \textbf{85.23} & \textbf{0.009} & \textbf{0.912} & \textbf{0.024} & \textbf{0.761} & \textbf{0.041} & \textbf{0.800} & 0.063 & 0.581 & \textbf{0.032} & \textbf{0.689} & \textbf{0.080} & \textbf{0.457} \\
\bottomrule
\end{tabular}%
}
\end{table*}

\section{Experiment}
We evaluate the proposed \textbf{HandTouch} framework on HT-Bench. To assess its representation capability, we compare it with representative tactile encoder baselines that are commonly used in tactile perception and manipulation tasks, including a CNN-based encoder~\citep{lee2026symmetry}, ResNet-18~\citep{Calandra2018more}, a VQ-VAE-based encoder~\citep{xu2025unit}, and a ViT-based encoder~\citep{zhao2024transferable}. For a fair comparison, all baselines are pretrained on the same training split and evaluated under the same HT-Bench protocols.

\subsection{Comparison with Baseline Encoders}
\label{sec:comparison}

We first compare HandTouch with representative tactile representation baselines on fine-grained tactile similarity retrieval, masked tactile inpainting, and vision-to-tactile synthesis. The results are summarized in \cref{tab:main_results}. Overall, HandTouch achieves the best performance on most evaluation metrics, demonstrating stronger tactile structural modeling, cross-modal alignment, and generalization under both in-distribution and OOD settings.

\textbf{Fine-Grained Tactile Similarity Retrieval.}
For fine-grained tactile similarity retrieval, HandTouch outperforms all baseline encoders. Hit@1 measures whether the SSIM-nearest tactile candidate is ranked first by the learned embedding similarity, while Recall@5 measures the fraction of SSIM top-5 candidates recovered in the top-5 retrieved candidates. Compared with the strongest baseline, the ViT-based encoder, HandTouch improves Hit@1 from 94.27\% to 99.27\% and Recall@5 from 74.65\% to 85.23\%. This indicates that the embedding space learned by HandTouch better preserves fine-grained structural similarity among tactile observations. In contrast, CNN- and ResNet-based models can capture local spatial patterns but are less effective in organizing tactile samples according to global structural correspondence. VQ-VAE performs notably worse on retrieval, suggesting that reconstruction-oriented discrete latent representations alone may lose subtle structural cues that are important for fine-grained tactile matching.

\textbf{Spatially Masked Tactile Inpainting.}
For masked tactile inpainting, HandTouch achieves the lowest reconstruction error and the highest contact overlap on the standard test split. Specifically, it obtains a full-map RMSE of 0.010 and a full-map cIoU of 0.911, substantially outperforming all baselines. We define contact IoU (cIoU) as:
\[\mathrm{cIoU} = \frac{\sum_{i,j} \min(P_{i,j}, \hat{P}_{i,j})}{\sum_{i,j} \max(P_{i,j}, \hat{P}_{i,j})}\]
where $P_{i,j}$ and $\hat{P}_{i,j}$ denote the ground-truth and predicted pressure values at pixel $(i,j)$, respectively. HandTouch also performs well in masked hole regions, achieving 0.024 RMSE and 0.761 cIoU, which demonstrates its ability to infer missing tactile responses from visible tactile context and learned spatial priors. Under the OOD setting, HandTouch maintains the best full-map reconstruction performance, with 0.041 RMSE and 0.800 cIoU. However, ResNet-18 performs better on OOD hole-region metrics. This suggests that although HandTouch generalizes well at the global tactile-map level, precise reconstruction of severely corrupted local regions in unseen tasks remains challenging.

\begin{figure*}[ht!]
    \centering
    \includegraphics[width=\linewidth]{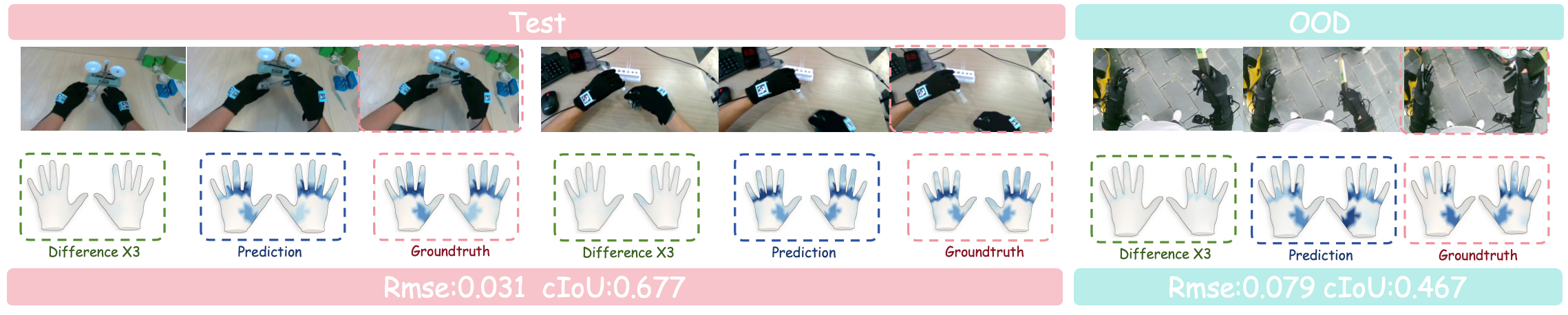}
    \caption{\textbf{Qualitative results of multimodal tactile frame prediction.}
    HandTouch predicts tactile distributions that closely match the ground truth under both in-distribution and OOD settings. Prediction errors are magnified by a factor of 3 for better visualization. These results illustrate the temporal contact modeling capability of HandTouch.}
    \label{fig:tactile_predict}
\end{figure*}

\textbf{Vision-to-Tactile Synthesis.}
For vision-to-tactile synthesis, HandTouch demonstrates strong cross-modal predictive capability. On the standard test split, HandTouch reduces the full-map RMSE to 0.032 and improves cIoU to 0.689, outperforming CNN-, ResNet-, VQ-VAE-, and ViT-based baselines. These results indicate that HandTouch learns effective vision--tactile correspondence, allowing it to infer tactile pressure distributions from visual observations. Under the OOD setting, HandTouch achieves the highest cIoU of 0.457 and 0.080 RMSE, indicating better cross-modal generalization to unseen tasks. 

\textbf{Multimodal Tactile Frame Prediction.}
Unlike the static evaluation tracks, multimodal tactile frame prediction requires temporal modeling from recent visual observations $\mathbf{v}{T-2}$ and past tactile trajectories $\mathbf{t}{T-2}$. Since the baselines are designed as single-frame tactile encoders, we report this temporal prediction track separately to evaluate the predictive capability of HandTouch. As shown in \cref{fig:tactile_predict}, HandTouch predicts tactile distributions $\mathbf{t}_T$ that closely match the ground truth. It achieves accurate prediction on the test split (RMSE: 0.031, cIoU: 0.677) and maintains reasonable performance under OOD tasks, demonstrating its ability to model temporal contact dynamics during continuous interaction.


\begin{table}[t]
\centering
\caption{Ablation study of HandTouch on masked tactile inpainting and vision-to-tactile synthesis. We report cIoU on the standard test and task-level OOD splits. H-cIoU is computed over masked hole regions for inpainting, while F-cIoU is computed over the full tactile map for RGB$\rightarrow$Tac. Higher is better, and the best result in each column is highlighted in bold.}
\label{tab:ablation}
\setlength{\tabcolsep}{3.2pt}
\small
\begin{tabular}{lcccc}
\toprule
\textbf{Variant} 
& \multicolumn{2}{c}{\textbf{Inpainting}} 
& \multicolumn{2}{c}{\textbf{RGB$\rightarrow$Tac}} \\
\cmidrule(lr){2-3} \cmidrule(lr){4-5}
& \shortstack{\textbf{H-cIoU}\\\textbf{Test}$\uparrow$}
& \shortstack{\textbf{H-cIoU}\\\textbf{OOD}$\uparrow$}
& \shortstack{\textbf{F-cIoU}\\\textbf{Test}$\uparrow$}
& \shortstack{\textbf{F-cIoU}\\\textbf{OOD}$\uparrow$} \\
\midrule
w/o drop     & 0.712 & 0.567 & 0.678 & 0.453 \\
w/o codebook & \textbf{0.765} & 0.529 & \textbf{0.708} & 0.454 \\
w/o vision   & 0.698 & 0.556 & 0.467 & 0.415 \\
Ours         & 0.761 & \textbf{0.581} & 0.689 & \textbf{0.457} \\
\bottomrule
\end{tabular}
\end{table}

\subsection{Ablation Study}

We conduct ablation studies to evaluate the contribution of key components in HandTouch, including the drop-based training strategy, the vector-quantized codebook, and visual conditioning. As shown in Table~\ref{tab:ablation}, the full model achieves the best OOD performance on both masked tactile inpainting and RGB$\rightarrow$Tac, demonstrating stronger generalization to unseen tasks.

Removing the drop-based strategy leads to consistent performance degradation across all metrics, indicating that tactile corruption regularization helps the model better infer missing contact patterns from incomplete observations. Removing the codebook slightly improves in-distribution performance, but significantly reduces OOD inpainting H-cIoU from 0.581 to 0.529. This suggests that the discrete codebook serves as an effective representation bottleneck, improving the transferability of tactile features rather than merely fitting the training distribution.

Visual conditioning is also essential. Without visual input, performance drops substantially, especially for RGB$\rightarrow$Tac, where test F-cIoU decreases from 0.689 to 0.467. This confirms that egocentric visual cues provide critical cross-modal information for tactile synthesis and also benefit masked tactile reconstruction. Overall, these results validate the proposed design: drop-based training improves robustness, the codebook enhances OOD generalization, and visual conditioning strengthens vision--touch alignment.



\begin{figure}[ht!]
    \centering
    \includegraphics[width=\linewidth]{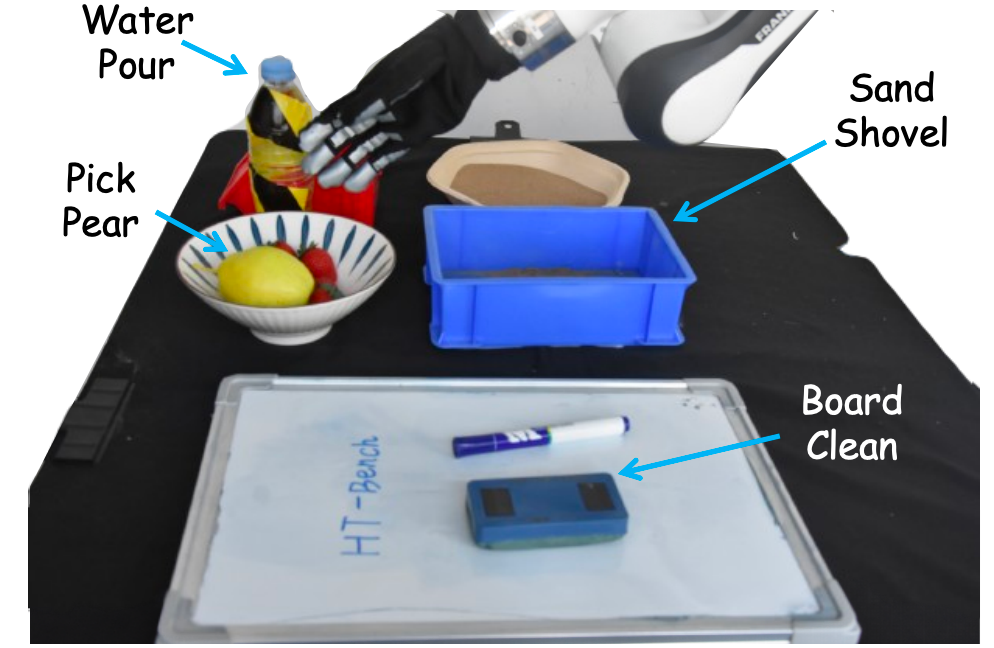}
    \caption{
    Real-world downstream evaluation setup. 
    We evaluate the learned tactile representation on four contact-rich manipulation tasks, including water pouring, pear picking, sand shoveling, and board cleaning.
    }
    \label{fig:real_setup}
\end{figure}


\begin{table}[th!] 
\centering \caption{Success rates (\%) on real-world downstream manipulation tasks. Each task is evaluated over 15 trials.} \label{tab:real_robot_results} \scriptsize \setlength{\tabcolsep}{2.4pt} \renewcommand{\arraystretch}{1.12} \begin{tabular}{@{}lccccc@{}} \toprule \textbf{Method} & \makecell{\textbf{Board}\\\textbf{Clean}} & \makecell{\textbf{Pear}\\\textbf{Pick}} & \makecell{\textbf{Water}\\\textbf{Pour}} & \makecell{\textbf{Sand}\\\textbf{Shovel}} & \textbf{Avg.} \\ \midrule ResNet-Scratch & 20.0 & 53.3 & 13.3 & 20.0 & 26.7 \\ ResNet-Trained & 46.7 & 73.3 & 33.3 & 46.7 & 50.0 \\ ViT & 40.0 & 73.3 & 33.3 & 33.3 & 45.0 \\ \addlinespace[1pt] \textbf{HandTouch} & \textbf{66.7} & \textbf{86.7} & \textbf{53.3} & \textbf{66.7} & \textbf{68.3} \\ \bottomrule \end{tabular} 
\end{table}

\subsection{Real-World Downstream Evaluation}

To further evaluate whether the learned tactile representations can transfer to real robotic manipulation(seen in fig. \ref{fig:real_setup}), we conduct downstream experiments on four contact-rich tasks: board cleaning, pear picking, water pouring, and sand shoveling. Each task is evaluated over 15 trials, and we report the success rate of each method in Table~\ref{tab:real_robot_results}.

Overall, HandTouch achieves the best performance across all tasks, obtaining an average success rate of 68.3\%, compared with 50.0\% for the trained ResNet encoder and 45.0\% for the ViT encoder. The improvement is consistent across different interaction types. On board cleaning and sand shoveling, where maintaining stable and distributed contact is critical, HandTouch improves the success rate to 66.7\%, outperforming the strongest baseline by 20.0 percentage points. On pear picking, HandTouch reaches 86.7\%, suggesting that the learned representation better captures fine-grained contact cues for stable grasping. Water pouring remains the most challenging task for all methods, as it requires precise contact and motion coordination; nevertheless, HandTouch still achieves the highest success rate of 53.3\%.

These results indicate that the proposed representation is not only effective on offline HT-Bench evaluations, but also provides practical benefits for real-world manipulation. In particular, the consistent gains over both convolutional and transformer-based baselines suggest that the progressive spatial, cross-modal, and temporal training design improves the transferability of tactile features to downstream contact-rich robotic tasks.

\section{Conclusion and Limitation}

In this paper, we introduce \textbf{HT-Bench}, a large-scale multi-task benchmark for dexterous tactile representation learning with egocentric vision and full-hand tactile data. HT-Bench evaluates tactile encoders on fine-grained tactile retrieval, masked tactile inpainting, vision-to-tactile synthesis, and multimodal tactile frame prediction, with task-level OOD splits for generalization assessment. We further propose \textbf{HandTouch}, a progressive vector-quantized vision--tactile encoder trained with spatial, cross-modal, and temporal objectives.

HandTouch consistently outperforms representative baselines on HT-Bench, demonstrating stronger contact-structure modeling, vision--touch alignment, temporal reasoning, and OOD generalization. Real-world downstream experiments on four contact-rich tasks further validate its transferability, where HandTouch achieves the best average success rate of 68.3\%, surpassing the strongest baseline by 18.3 percentage points. These results show that egocentric full-hand tactile data offers a scalable foundation for general tactile representation learning in dexterous manipulation.

A limitation of HT-Bench is that it focuses on egocentric vision paired with full-hand tactile sensing, without yet covering broader sensing paradigms such as fingertip optical tactile sensors, force/torque sensors, skin-like taxel arrays, or non-hand embodiments. Extending the benchmark to more diverse hardware platforms, sensing modalities, and robotic embodiments is an important future direction.

\bibliographystyle{aaai}
\bibliography{references}

\begin{thebibliography}{}

\bibitem[\protect\citeauthoryear{Bao \bgroup et al\mbox.\egroup }{2021}]{bao2021beit}
Bao, H.; Dong, L.; Piao, S.; and Wei, F.
\newblock 2021.
\newblock Beit: Bert pre-training of image transformers.
\newblock {\em arXiv preprint arXiv:2106.08254}.

\bibitem[\protect\citeauthoryear{Calandra \bgroup et al\mbox.\egroup }{2018}]{Calandra2018more}
Calandra, R.; Owens, A.; Jayaraman, D.; Lin, J.; Yuan, W.; Malik, J.; Adelson, E.~H.; and Levine, S.
\newblock 2018.
\newblock More than a feeling: Learning to grasp and regrasp using vision and touch.
\newblock {\em IEEE Robotics and Automation Letters} 3(4):3300--3307.

\bibitem[\protect\citeauthoryear{Cao \bgroup et al\mbox.\egroup }{2026}]{cao2026tactile}
Cao, Z.; Tian, D.; Guan, R.; Mu, Y.; Sun, X.; Liang, S.; Liu, D.; Huang, T.; Yue, Y.; Ding, H.; Fang, B.; Zhou, A.; Han, Q.-L.; and Xiong, H.
\newblock 2026.
\newblock Tactile-based multimodal fusion in embodied intelligence: A survey of vision, language, and contact-driven paradigms.

\bibitem[\protect\citeauthoryear{Chen \bgroup et al\mbox.\egroup }{2026a}]{chen2026univtac}
Chen, B.; Wan, W.; Chen, T.; Guo, X.; Xu, C.; Qi, Y.; Zhang, H.; Wu, L.; Xu, T.; Li, Z.; Wu, Y.; Li, R.; Yang, X.; Luo, P.; Sui, W.; and Mu, Y.
\newblock 2026a.
\newblock Univtac: A unified simulation platform for visuo-tactile manipulation data generation, learning, and benchmarking.

\bibitem[\protect\citeauthoryear{Chen \bgroup et al\mbox.\egroup }{2026b}]{chen2026multimodal}
Chen, H.; Xu, J.; Chen, H.; Hong, K.; Huang, B.; Liu, C.; Mao, J.; Li, Y.; Du, Y.; and Driggs-Campbell, K.
\newblock 2026b.
\newblock Multi-modal manipulation via multi-modal policy consensus.

\bibitem[\protect\citeauthoryear{Chi \bgroup et al\mbox.\egroup }{2024}]{chi2024multi}
Chi, H.-G.; Barreiros, J.; Mercat, J.; Ramani, K.; and Kollar, T.
\newblock 2024.
\newblock Multi-modal representation learning with tactile data.
\newblock In {\em 2024 IEEE/RSJ International Conference on Intelligent Robots and Systems (IROS)},  9660--9667.
\newblock IEEE.

\bibitem[\protect\citeauthoryear{Cytowic}{2002}]{cytowic2002synesthesia}
Cytowic, R.~E.
\newblock 2002.
\newblock {\em Synesthesia: A union of the senses}.
\newblock MIT press.

\bibitem[\protect\citeauthoryear{Dosovitskiy \bgroup et al\mbox.\egroup }{2021}]{dosovitskiy2021vit}
Dosovitskiy, A.; Beyer, L.; Kolesnikov, A.; Weissenborn, D.; Zhai, X.; Unterthiner, T.; Dehghani, M.; Minderer, M.; Heigold, G.; Gelly, S.; Uszkoreit, J.; and Houlsby, N.
\newblock 2021.
\newblock An image is worth 16x16 words: Transformers for image recognition at scale.

\bibitem[\protect\citeauthoryear{Fang \bgroup et al\mbox.\egroup }{2026}]{fang2026unidextok}
Fang, D.; Wu, Y.; Zhong, Y.; Zhang, R.; Wang, Y.; Jia, X.; and Jiang, Y.-G.
\newblock 2026.
\newblock Unidextok: A unified dexterous hand tokenizer from real data.
\newblock {\em arXiv preprint arXiv:2606.10683}.

\bibitem[\protect\citeauthoryear{Feng \bgroup et al\mbox.\egroup }{2025}]{feng2025anytouch}
Feng, R.; Hu, J.; Xia, W.; Gao, T.; Shen, A.; Sun, Y.; Fang, B.; and Hu, D.
\newblock 2025.
\newblock Anytouch: Learning unified static-dynamic representation across multiple visuo-tactile sensors.
\newblock {\em arXiv preprint arXiv:2502.12191}.

\bibitem[\protect\citeauthoryear{Feng \bgroup et al\mbox.\egroup }{2026}]{feng2026anytouch}
Feng, R.; Zhou, Y.; Mei, S.; Zhou, D.; Wang, P.; Cui, S.; Fang, B.; Yao, G.; and Hu, D.
\newblock 2026.
\newblock Anytouch 2: General optical tactile representation learning for dynamic tactile perception.
\newblock {\em arXiv preprint arXiv:2602.09617}.

\bibitem[\protect\citeauthoryear{Higuera \bgroup et al\mbox.\egroup }{2024}]{higuera2024sparsh}
Higuera, C.; Sharma, A.; Bodduluri, C.~K.; Fan, T.; Lancaster, P.; Kalakrishnan, M.; Kaess, M.; Boots, B.; Lambeta, M.; Wu, T.; and Mukadam, M.
\newblock 2024.
\newblock Sparsh: Self-supervised touch representations for vision-based tactile sensing.

\bibitem[\protect\citeauthoryear{Huang \bgroup et al\mbox.\egroup }{2026}]{huang2026tactile}
Huang, Y.; Lin, P.; Li, W.; Li, D.; Li, J.; Jiang, J.; Xiao, C.; and Jiao, Z.
\newblock 2026.
\newblock Tactile-force alignment in vision-language-action models for force-aware manipulation.
\newblock {\em arXiv preprint arXiv:2601.20321}.

\bibitem[\protect\citeauthoryear{Lee, Grimaldi, and Yu}{2026}]{lee2026symmetry}
Lee, W.; Grimaldi, M.; and Yu, T.
\newblock 2026.
\newblock Symmetry-aware fusion of vision and tactile sensing via bilateral force priors for robotic manipulation.
\newblock {\em arXiv preprint arXiv:2602.13689}.

\bibitem[\protect\citeauthoryear{Li \bgroup et al\mbox.\egroup }{2025}]{li2025visuo}
Li, Y.; Jin, Z.; Liu, J.; and Ma, D.
\newblock 2025.
\newblock Visuo-tactile feedback policies for terminal assembly facilitated by reinforcement learning.
\newblock {\em Frontiers in Robotics and AI} 12:1660244.

\bibitem[\protect\citeauthoryear{Li \bgroup et al\mbox.\egroup }{2026}]{li2026simultaneous}
Li, Y.; Chen, Y.; Zhao, Z.; Li, P.; Liu, T.; Huang, S.; and Zhu, Y.
\newblock 2026.
\newblock Simultaneous tactile-visual perception for learning multimodal robot manipulation.

\bibitem[\protect\citeauthoryear{Lin \bgroup et al\mbox.\egroup }{2015}]{lin2015microsoftcoco}
Lin, T.-Y.; Maire, M.; Belongie, S.; Bourdev, L.; Girshick, R.; Hays, J.; Perona, P.; Ramanan, D.; Zitnick, C.~L.; and Dollár, P.
\newblock 2015.
\newblock Microsoft coco: Common objects in context.

\bibitem[\protect\citeauthoryear{Lin \bgroup et al\mbox.\egroup }{2025}]{lin2025pp}
Lin, P.; Huang, Y.; Li, W.; Ma, J.; Xiao, C.; and Jiao, Z.
\newblock 2025.
\newblock Pp-tac: Paper picking using tactile feedback in dexterous robotic hands.
\newblock {\em arXiv preprint arXiv:2504.16649}.

\bibitem[\protect\citeauthoryear{Liu \bgroup et al\mbox.\egroup }{2024}]{liu2024masked}
Liu, Q.; Ye, Q.; Sun, Z.; Cui, Y.; Li, G.; and Chen, J.
\newblock 2024.
\newblock Masked visual-tactile pre-training for robot manipulation.
\newblock In {\em 2024 IEEE International Conference on Robotics and Automation (ICRA)},  13859--13875.
\newblock IEEE.

\bibitem[\protect\citeauthoryear{Liu \bgroup et al\mbox.\egroup }{2025}]{liu2025vtdexmanip}
Liu, Q.; Cui, Y.; Sun, Z.; Li, G.; Chen, J.; and Ye, Q.
\newblock 2025.
\newblock Vtdexmanip: A dataset and benchmark for visual-tactile pretraining and dexterous manipulation with reinforcement learning.
\newblock In {\em The Thirteenth International Conference on Learning Representations}.

\bibitem[\protect\citeauthoryear{Luo \bgroup et al\mbox.\egroup }{2026}]{luo2026omniumi}
Luo, S.; Li, Y.; Hu, Y.; Yu, C.; Xu, C.; Zhang, J.; Yao, G.; Huang, T.; He, R.; and Wang, Z.
\newblock 2026.
\newblock Omniumi: Towards physically grounded robot learning via human-aligned multimodal interaction.

\bibitem[\protect\citeauthoryear{Pei \bgroup et al\mbox.\egroup }{2026}]{DexMove_ICLR2026}
Pei, L.; Yuzhe, H.; Wanlin, L.; Chenxi, X.; and Ziyuan, J.
\newblock 2026.
\newblock Dexmove: Learning tactile-guided non-prehensile manipulation with dexterous hands.
\newblock In {\em The Fourteenth International Conference on Learning Representations}.

\bibitem[\protect\citeauthoryear{Radford \bgroup et al\mbox.\egroup }{2021}]{radford2021clip}
Radford, A.; Kim, J.~W.; Hallacy, C.; Ramesh, A.; Goh, G.; Agarwal, S.; Sastry, G.; Askell, A.; Mishkin, P.; Clark, J.; Krueger, G.; and Sutskever, I.
\newblock 2021.
\newblock Learning transferable visual models from natural language supervision.

\bibitem[\protect\citeauthoryear{Razavi, Van~den Oord, and Vinyals}{2019}]{razavi2019generating}
Razavi, A.; Van~den Oord, A.; and Vinyals, O.
\newblock 2019.
\newblock Generating diverse high-fidelity images with vq-vae-2.
\newblock {\em Advances in neural information processing systems} 32.

\bibitem[\protect\citeauthoryear{Russakovsky \bgroup et al\mbox.\egroup }{2015}]{russakovsky2015imagenet}
Russakovsky, O.; Deng, J.; Su, H.; Krause, J.; Satheesh, S.; Ma, S.; Huang, Z.; Karpathy, A.; Khosla, A.; Bernstein, M.; Berg, A.~C.; and Fei-Fei, L.
\newblock 2015.
\newblock Imagenet large scale visual recognition challenge.

\bibitem[\protect\citeauthoryear{Schneider \bgroup et al\mbox.\egroup }{2025}]{schneider2025tactile}
Schneider, T.; Duret, G.; de~Farias, C.; Calandra, R.; Chen, L.; and Peters, J.
\newblock 2025.
\newblock Tactile mnist: Benchmarking active tactile perception.

\bibitem[\protect\citeauthoryear{Song \bgroup et al\mbox.\egroup }{2025}]{song2025opentouch}
Song, Y.~R.; Li, J.; Fu, R.; Murphy, D.; Zhou, K.; Shiv, R.; Li, Y.; Xiong, H.; Owens, C.~E.; Du, Y.; Luo, Y.; Cheng, X.; Torralba, A.; Matusik, W.; and Liang, P.~P.
\newblock 2025.
\newblock Opentouch: Bringing full-hand touch to real-world interaction.

\bibitem[\protect\citeauthoryear{Van Den~Oord, Vinyals, and others}{2017}]{van2017neural}
Van Den~Oord, A.; Vinyals, O.; et~al.
\newblock 2017.
\newblock Neural discrete representation learning.
\newblock {\em Advances in neural information processing systems} 30.

\bibitem[\protect\citeauthoryear{Wang \bgroup et al\mbox.\egroup }{2026}]{humanego2026}
Wang, Z.; He, B.; Yu, K.; Lee, S.; Gao, R.; Huang, F.; and Aloimonos, Y.
\newblock 2026.
\newblock Humanego: Zero-shot robot learning from minutes of human egocentric videos.

\bibitem[\protect\citeauthoryear{Wu \bgroup et al\mbox.\egroup }{2026}]{wu2026dexgrasp}
Wu, Y.; Lin, Y.; Lao, W.; Lin, Y.; Wei, Y.-L.; Zheng, W.-S.; and Wu, A.
\newblock 2026.
\newblock Dexgrasp-zero: A morphology-aligned policy for zero-shot cross-embodiment dexterous grasping.
\newblock {\em arXiv preprint arXiv:2603.16806}.

\bibitem[\protect\citeauthoryear{Xie \bgroup et al\mbox.\egroup }{2025}]{xie2025universal}
Xie, Y.; Li, M.; Li, S.; Li, X.; Chen, G.; Ma, F.; Yu, F.~R.; and Ding, W.
\newblock 2025.
\newblock Universal visuo-tactile video understanding for embodied interaction.

\bibitem[\protect\citeauthoryear{Xie \bgroup et al\mbox.\egroup }{2026}]{xie2026universal}
Xie, Y.; Li, M.; Li, S.; Li, X.; Chen, G.; Ma, F.; Yu, F.; and Ding, W.
\newblock 2026.
\newblock Universal visuo-tactile video understanding for embodied interaction.
\newblock {\em Advances in Neural Information Processing Systems} 38:127864--127883.

\bibitem[\protect\citeauthoryear{Xu \bgroup et al\mbox.\egroup }{2025}]{xu2025unit}
Xu, Z.; Uppuluri, R.; Zhang, X.; Fitch, C.; Crandall, P.~G.; Shou, W.; Wang, D.; and She, Y.
\newblock 2025.
\newblock Unit: Data efficient tactile representation with generalization to unseen objects.

\bibitem[\protect\citeauthoryear{Yang \bgroup et al\mbox.\egroup }{2024a}]{yang2024bindingtouch}
Yang, F.; Feng, C.; Chen, Z.; Park, H.; Wang, D.; Dou, Y.; Zeng, Z.; Chen, X.; Gangopadhyay, R.; Owens, A.; and Wong, A.
\newblock 2024a.
\newblock Binding touch to everything: Learning unified multimodal tactile representations.

\bibitem[\protect\citeauthoryear{Yang \bgroup et al\mbox.\egroup }{2024b}]{yang2024binding}
Yang, F.; Feng, C.; Chen, Z.; Park, H.; Wang, D.; Dou, Y.; Zeng, Z.; Chen, X.; Gangopadhyay, R.; Owens, A.; and Wong, A.
\newblock 2024b.
\newblock Binding touch to everything: Learning unified multimodal tactile representations.
\newblock In {\em 2024 IEEE/CVF Conference on Computer Vision and Pattern Recognition (CVPR)},  26330--26343.

\bibitem[\protect\citeauthoryear{Zhai \bgroup et al\mbox.\egroup }{2023}]{zhai2023sigmoid}
Zhai, X.; Mustafa, B.; Kolesnikov, A.; and Beyer, L.
\newblock 2023.
\newblock Sigmoid loss for language image pre-training.

\bibitem[\protect\citeauthoryear{Zhao \bgroup et al\mbox.\egroup }{2024}]{zhao2024transferable}
Zhao, J.; Ma, Y.; Wang, L.; and Adelson, E.~H.
\newblock 2024.
\newblock Transferable tactile transformers for representation learning across diverse sensors and tasks.
\newblock {\em arXiv preprint arXiv:2406.13640}.

\bibitem[\protect\citeauthoryear{Zheng \bgroup et al\mbox.\egroup }{2026}]{zheng2026egoscalescalingdexterousmanipulation}
Zheng, R.; Niu, D.; Xie, Y.; Wang, J.; Xu, M.; Jiang, Y.; Castañeda, F.; Hu, F.; Tan, Y.~L.; Fu, L.; Darrell, T.; Huang, F.; Zhu, Y.; Xu, D.; and Fan, L.
\newblock 2026.
\newblock Egoscale: Scaling dexterous manipulation with diverse egocentric human data.

\bibitem[\protect\citeauthoryear{Zhou \bgroup et al\mbox.\egroup }{2026}]{zhou2026touchanything}
Zhou, J.; Gao, Z.; Hong, F.; Liu, Z.; Zhang, G.; Dai, W.; Zhen, R.; Lyu, C.; Wu, H.; Mao, Y.; Wang, X.; Jiang, Y.; Ding, W.; and Yang, S.
\newblock 2026.
\newblock Touchanything: A dataset and framework for bimanual tactile estimation from egocentric video.

\bibitem[\protect\citeauthoryear{Zorin \bgroup et al\mbox.\egroup }{2026}]{zorin2026taco}
Zorin, A.; Si, Z.; Park, M.; Park, J.; Buynitsky, A.; Bhadang, S.; Park, T.; Yoon, S.~J.; Park, Y.-L.; Kroemer, O.; et~al.
\newblock 2026.
\newblock Taco: Benchmarking tactile sensors for object manipulation.
\newblock {\em arXiv preprint arXiv:2605.21976}.

\end{thebibliography}
\end{document}